\documentclass[dvipsnames]{article} %
\usepackage{colm2024_conference}

\usepackage{booktabs}
\usepackage{enumitem}
\usepackage{wrapfig}
\usepackage{algorithm}
\usepackage{algpseudocode}
\usepackage{graphicx}
\usepackage[misc]{ifsym}
\usepackage{multicol} 
\usepackage{microtype}
\usepackage{amsmath}
\usepackage{colortbl}
\usepackage[utf8]{inputenc}
\usepackage[T1]{fontenc}
\definecolor{lightgray}{rgb}{0.9,0.9,0.9}
\usepackage{caption}
\usepackage{subcaption}
\usepackage{graphicx}
\usepackage{subcaption}
\usepackage{setspace}
\usepackage{url}
\usepackage{multirow}
\usepackage{tabularx}
\usepackage{blindtext}
\usepackage{pgfplots}
\pgfplotsset{compat=1.18} 
\usepackage{tikz}
\usetikzlibrary{er,positioning,bayesnet}
\usepackage{makecell}
\usepackage{tipa}
\usepackage{siunitx}
\usepackage{nicefrac}
\usepackage{listings}
\usepackage[raster,skins, most]{tcolorbox} %
\usepackage{xltabular}
\usepackage{adjustbox}
\usepackage{xurl}
\usepackage{rotating}
\usepackage[normalem]{ulem}

\usepackage{fontawesome}

\usepackage[table]{xcolor}

\useunder{\uline}{\ul}{}

\usepackage{amsmath,amsfonts,bm}

\def\eqref#1{equation~\ref{#1}}
\def\1{\bm{1}}

\DeclareMathAlphabet{\mathsfit}{\encodingdefault}{\sfdefault}{m}{sl}
\SetMathAlphabet{\mathsfit}{bold}{\encodingdefault}{\sfdefault}{bx}{n}

\renewcommand{\ghlink}{https://github.com/Alibaba-NLP/DeepResearch}

\newcommand*\justify{%
  \fontdimen2\font=0.4em% interword space
  \fontdimen3\font=0.2em% interword stretch
  \fontdimen4\font=0.1em% interword shrink
  \fontdimen7\font=0.1em% extra space
  \hyphenchar\font=`\-% allowing hyphenation
}

\renewcommand{\texttt}[1]{%
  \begingroup
  \ttfamily
  \begingroup\lccode`~=`/\lowercase{\endgroup\def~}{/\discretionary{}{}{}}%
  \begingroup\lccode`~=`[\lowercase{\endgroup\def~}{[\discretionary{}{}{}}%
  \begingroup\lccode`~=`.\lowercase{\endgroup\def~}{.\discretionary{}{}{}}%
  \catcode`/=\active\catcode`[=\active\catcode`.=\active
  \justify\scantokens{#1\noexpand}%
  \endgroup
}

\usepackage{makecell}
\usetikzlibrary{tikzmark}
\makeatletter
\newcommand*\myfontsize{%
  \@setfontsize\myfontsize{7}{8}%
}
\makeatother

\definecolor{uclablue}{RGB}{159, 195, 224}

\definecolor{uclagold}{RGB}{255, 240, 180}

\definecolor{aliceblue}{RGB}{255, 238, 241}

\definecolor{cadmiumgreen}{rgb}{0.0, 0.42, 0.24}

\definecolor{myred}{rgb}{0.7, 0.3, 0.0}
\definecolor{myblue}{rgb}{0.2, 0.3, 0.6}
\definecolor{babygreen}{rgb}{0.85, 0.97, 0.85}

\definecolor{purple1}{RGB}{126, 107, 196}
\definecolor{purple2}{RGB}{199, 158, 207}
\definecolor{purple3}{RGB}{214, 200, 255}
\definecolor{purple4}{RGB}{254, 240, 255}

\definecolor{deepblue}{RGB}{48, 58, 82}

\newcommand{\symboletongyi}{\raisebox{0pt}{~\includegraphics[scale=0.012]{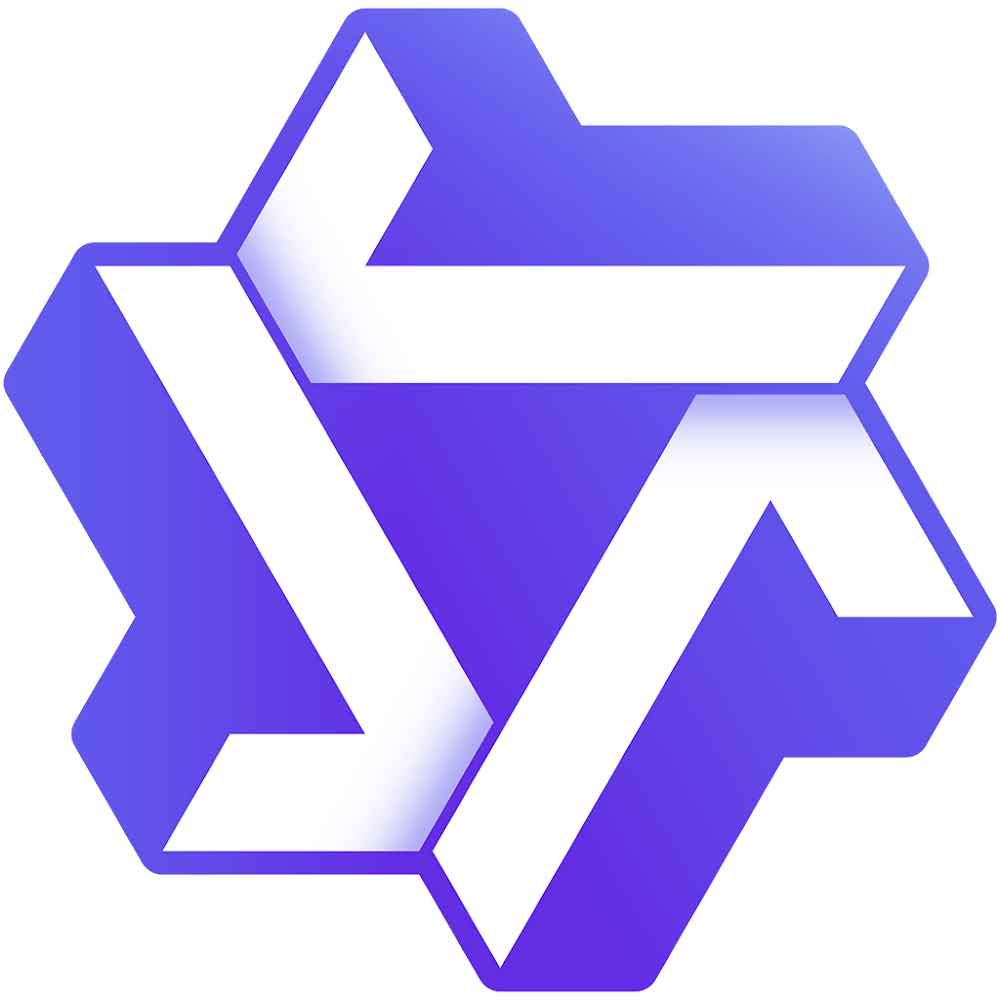}}~}

\definecolor{deepPurple}{HTML}{330066}
\definecolor{uclablue_old}{rgb}{0.15, 0.45, 0.68}
\hypersetup{
    breaklinks,
    citecolor=uclablue_old,
    colorlinks=true,
}

\newtcolorbox{mybox}[2][]
  {colback = black!5!white, colframe = black!75!black, fonttitle = \bfseries,
    colbacktitle = black!100!black, enhanced, before upper={\fontsize{8}{11}\obeyspaces\obeylines\selectfont}, fontupper=\selectfont,
    attach boxed title to top left={yshift=-2.2mm,xshift=4mm},
    title=#2,#1}

\title{%
\begin{tabular}[t]{l} % 注意这里是 c，不是 l
  \parbox[t]{0.8\textwidth}{\centering % 加了\centering实现居中
    Tongyi DeepResearch Technical Report
  }
\end{tabular}
}

\author{%
\large Tongyi DeepResearch Team\thanks{Full author list available in the \hyperref[sec:contribution]{Contributions} section.}%
  \\[1em]               % ← 在上一行末尾插入 1 em 的竖直间距，相当于“空一行”
  {\fontsize{10pt}{11pt}\selectfont          % \large 放大字号；\bfseries 视需要加粗
Tongyi Lab\symboletongyi, Alibaba Group}\\
}

\begin{document}

\maketitle

\begin{abstract}
We present \textbf{Tongyi DeepResearch}, an agentic large language model, which is specifically designed for long-horizon, deep information-seeking research tasks.
To incentivize autonomous deep research agency, Tongyi DeepResearch is developed through an end-to-end training framework that combines agentic mid-training and agentic post-training, enabling scalable reasoning and information seeking across complex tasks.
We design a highly scalable data synthesis pipeline that is fully automatic, without relying on costly human annotation, and empowers all training stages.
By constructing customized environments for each stage, our system enables stable and consistent interactions throughout.
Tongyi DeepResearch, featuring 30.5 billion total parameters, with only 3.3 billion activated per token, achieves state-of-the-art performance across a range of agentic deep research benchmarks, including Humanity's Last Exam, BrowseComp, BrowseComp-ZH, WebWalkerQA, xbench-DeepSearch, FRAMES and xbench-DeepSearch-2510.
We open-source the model, framework, and complete solutions to empower the community. 

\end{abstract}

\begin{figure}[h]
    \centering
    \includegraphics[width=1.0\linewidth]{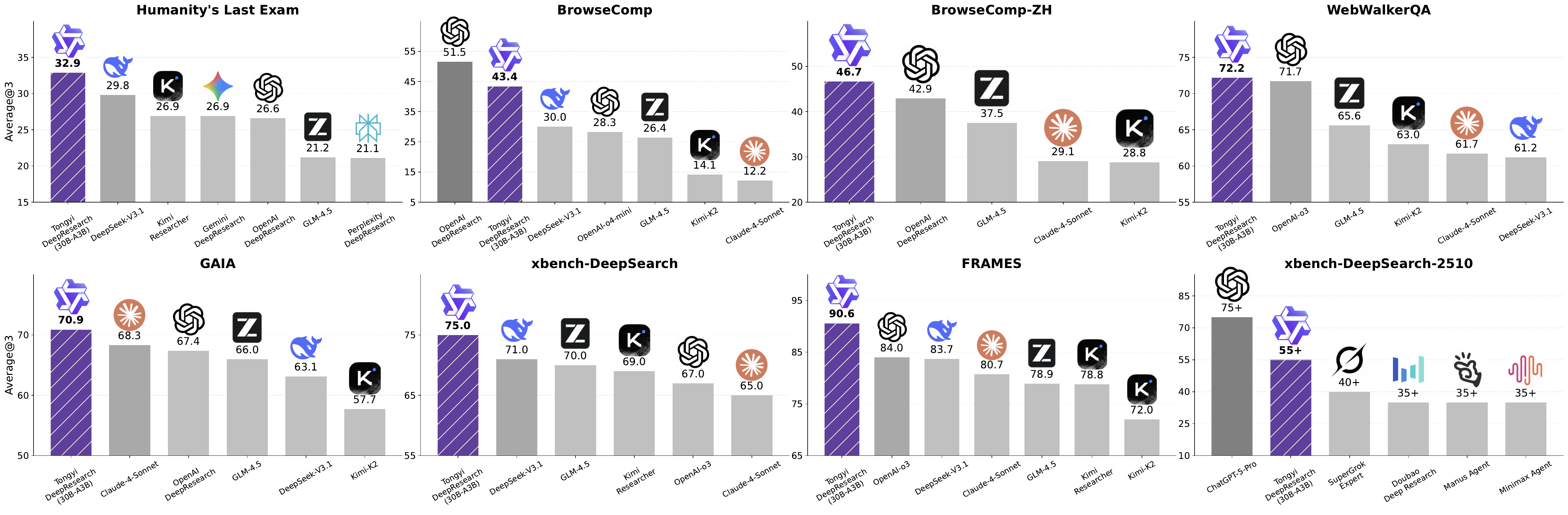}
    \caption{Benchmark performance of Tongyi DeepResearch.}
    \label{fig:abs_fig}
\end{figure}
\vfill

\section{Introduction}

\label{sec:intro}
As we advance toward Artificial General Intelligence (AGI), the emergence of Deep Research agents offers a promising paradigm for augmenting and potentially liberating human intellectual productivity.
Deep research is a new agentic capability that autonomously conducts multi-step reasoning and information seeking on the internet for complex research tasks. 
It can be completed in tens of minutes, which would otherwise require several hours for a human~\citep{dr,clauderesearch,groksearch,geminiresearch}.
However, most deep research systems remain closed-source, and their intermediate research processes are inaccessible. 
While the community has made preliminary explorations in this area~\citep{wu2025webdancer,li2025websailor,tao2025webshaper}, there is still a lack of a systematic methodology and publicly available models that can be fully open-sourced and shared across the community.

In this work, we introduce \textbf{Tongyi DeepResearch}, opening the era of open-source AI researchers.
Our goal is to endow large language models (LLMs) with autonomous research capabilities agency, the ability to plan, search, reason, and synthesize knowledge across extended sequences of actions and diverse information sources.

Tongyi DeepResearch delivers several key advancements: 
\begin{itemize}[leftmargin=1.5em, itemsep=2pt, topsep=1pt, parsep=0pt]
    \item We propose an \textbf{end-to-end agentic training paradigm} that unifies agentic mid-training and agentic post-training, forming a scalable foundation for deep reasoning and information-seeking behaviors.
    Agentic mid-training cultivates inherent agentic biases by exposing the model to large-scale, high-quality agentic data, serving as a progressive transition from pre-training to post-training stages.
    Agentic post-training further unlocks the model’s potential via scalable multi-turn reinforcement learning on a strong base model.
    Together, they enable the model to gradually develop from basic interaction skills to advanced autonomous research behaviors. 
    \item  We design a \textbf{fully automated, highly scalable data synthesis pipeline} that eliminates human annotation while generating diverse, high-quality agent trajectories.
    We design stage-specific data synthesis strategies tailored to the objectives of each training phase, ensuring that every stage is supported by appropriately structured and targeted data.
    Synthetic data is highly scalable, fast to validate, and enables the construction of super-human-level datasets with stable distributions.
    It serves as an indispensable engine for agent training.
    \item We construct \textbf{stage-specific, customized environments} that rely on robust infrastructure to deliver consistent interactions for data synthesis across training stages.
    These environments allow the agent to engage in rich, specialized interactions that are tightly aligned with its developmental stage.
    They can take various forms, from prior world models to simulated environments and real-world interactive contexts.
\end{itemize}

Tongyi DeepResearch establishes a new state-of-the-art with substantially fewer parameters, comprising a total of 30.5 billion parameters while activating only 3.3 billion per token, building upon the Qwen3-30B-A3B-Base model~\citep{yang2025qwen3}.
Empirical evaluations on deep research benchmarks demonstrate the effectiveness of our agent.
Tongyi DeepResearch reaches 32.9 on Humanity's Last Exam, 43.4 on BrowseComp, 46.7 on BrowseComp-ZH, 72.2 on WebWalkerQA, 70.9 on GAIA, 75.0 on xbench-DeepSearch, 90.6 on FRAMES and 55.0 on xbench-DeepSearch-2510, outperforming strong baselines such as OpenAI-o3~\citep{o3} and Deepseek-V3.1~\citep{deepseekv3.1}.
We also provide a systematic analysis covering agentic reinforcement learning, synthetic data, offering key insights into the development of deep research agent.
In addition, we present the performance of Tongyi DeepResearch on general benchmarks, including AIME25, HMMT25 and SimpleQA. 
We believe that agentic models represent an emerging trend for the future, as models increasingly internalize agent-like capabilities and can autonomously invoke the appropriate tools to solve a wide range of problems.

In the following sections, we first outline the design principles underlying Tongyi DeepResearch.
We then describe the training pipeline, followed by a comprehensive evaluation of its performance. 
We release the model, framework, and end-to-end solutions to support and accelerate community research.
This technical report summarizes our main insights and aims to inspire further progress toward scalable and capable agentic systems.
\section{Design Principle}
\textbf{Agent Training Pipeline.} Agent training is inherently more complex and challenging than conventional LLM training. 
We introduce two stages in our agent training pipeline: mid-training and post-training. 
We integrate mid-training directly into the deep research training process, and co-design the end-to-end on-policy reinforcement learning algorithm and its underlying infrastructure for seamless scalability and stability. 
While most work only applies post-training phase for DeepResearch agents, we novelly introduce mid-training for agentic learning. 
General foundation models usually lack agentic inductive bias. 
Most general foundation models are typically pretrained on plain text crawled from the internet and then post-trained on instruction-following data. 
These datasets lack research-level questions and agentic behaviors, resulting in the model learns agentic capabilities and alignment simultaneously during the post-training phase. 
Agentic post-training on these general foundation models can result in sub-optimal outcomes and inherent optimization conflicts. 
Mid-training endows the pre-trained base model with substantial agentic prior knowledge, thereby bridging the gap between pretraining and agentic post-training. Mid-training phase provides a powerful \textbf{agentic foundation model} to support effective agentic post-training. 
During post-training, the model further internalizes deep research capabilities through reinforcement learning with supervised fine-tuning (SFT) for cold start. 
SFT teaches the model to reliably imitate curated demonstrations, establishing a stable behavioral baseline for research workflows and tool use. However, behavior cloning alone tends to produce mimicry without exploration. 
RL closes the loop with the environment, using reward signals to refine policies and to internalize agentic planning and execution.
In particular, reinforcement learning (1) explores optimal strategies through active interaction with the environment; (2) internalizes goal-directed planning and execution capabilities; and 3) achieves superior sample efficiency by prioritizing high-reward behaviors. 
The agent first acquires general agentic pattern during supervised fine-tuning phase, while reinforcement learning phase effectively pushes the limits of its agentic performance.

\textbf{Synthetic Data Centric Scaling.} Data serves as the foundation of training, while collecting data for DeepResearch problems is extremely hard. 
Deep research problems require agents' capability of connecting information, reasoning across sources and validating conclusions. 
Unlike pre-training data, which is naturally abundant, and conventional LLM post-training data, which is relatively easy to annotate, agentic data is inherently scarce. 
Research-level problems are difficult to obtain through natural texts from the web. 
Manually annotating these problems and agentic trajectories is extremely time-consuming and costly~\citep{bc_en}.
Building on the aforementioned agent training pipeline, agentic mid-training requires large-scale, diverse trajectories to align subsequent agent behaviors, while agentic post-training depends on high-quality, verifiable data to provide reliable reward signals. 
As a result, it is hard to rely on natural data to scale DeepResearch capability. 
Therefore, we focus on synthetic data with large language models.
Synthetic data contains several advantages over human annotations below:

\begin{itemize}[leftmargin=1.5em, itemsep=2pt, topsep=1pt, parsep=0pt]
    \item \textbf{Synthesizing research-level questions is easy to scale.} We can use LLMs to synthesize question-answer pair efficiently compared to manually annotating.
    \item \textbf{The pattern and diversity are easy to generalize.} LLMs are easy to understand the structure of hard problems and usually have rare insight into diverse patterns, while training annotators to understand the structure and patterns for research-level problems is time-consuming.
    \item \textbf{Synthesized data enables targeted meta-capability enhancement.} By decomposing complex agent tasks into fundamental meta-capabilities (\textit{e.g.}, planning, information synthesis, memory management), we can generate synthetic data that specifically targets and strengthens individual agent skills. 
    \item \textbf{Synthesized data can be verified easily.} It is much easier than finding the solution to the question, which is essential in human annotating. 
    \item \textbf{Synthesized data can provide data flywheels in training stages.} After one round of the agentic training pipeline, the trained agentic model can generate synthesized data with stronger reasoning and planning patterns. Data flywheel makes the agentic model evolve iteratively.
\end{itemize}
Based on these insights, we believe synthetic agentic data becomes the key to scaling deep-research agents. The synthetic data in all phases of the agentic training pipeline are designed in three steps: (1) synthesizing research-level questions; (2) Generating agentic behavior data; (3) Utilizing agentic data in training pipeline.

\textbf{Learning Through Environmental Interaction.} Environmental interaction plays a crucial role in agent intelligence emergence~\citep{silver2025welcome}. 
However, relying solely on real-world environments for the whole agent training stage faces fundamental challenges: \textbf{(1) Non-stationarity}. The dynamic nature of environments causes continuous distribution shift in training data, undermining learning stability; \textbf{(2) Interaction cost}. The tangible expense of each API call makes large-scale exploration economically prohibitive. These barriers render agent capability acquisition from the real world alone a formidable endeavor.

In Tongyi DeepResearch, we propose a fundamental reframing: \textit{environments should not be passively viewed as external reality, but actively designed as systems deeply coupled with the training process}. Specifically, we model environments into three forms, each striking a distinct balance between stability, fidelity, and cost:

\begin{itemize}[leftmargin=1.5em, itemsep=2pt, topsep=1pt, parsep=0pt]
    \item \textbf{Prior World Environment.} This environment provides task elements, tools, and state definitions, allowing agents to autonomously mine interaction trajectories based on pretrained knowledge without receiving actual environmental responses. 
    It offers perfect stability, zero interaction cost, and unlimited scalability, but lacks real-world feedback signals.
    \item \textbf{Simulated Environment.} This environment constructs controlled, reproducible replicas of real-world interactions locally. 
    It provides stability, rapid response, and low cost, enabling fast iteration and causal attribution analysis. 
    However, its data coverage is inherently limited, exhibiting a notable sim-to-real gap.
    \item \textbf{Real-world Environment.} This environment delivers the most authentic data distribution and feedback signals, serving as the ultimate proving ground for agent capabilities. Its advantage lies in absolute distributional fidelity; the cost is expensive interactions, significant non-stationarity, and exploration risks.
\end{itemize}

Building on this environmental insight, we adopt adaptive strategies for synthetic data generation and training. Specifically, (1) During agentic mid-training, we primarily leverage the Prior World Environment and Simulated Environment to generate large-scale synthetic data at minimal cost, ensuring efficient agentic ability bootstrapping; (2) During agentic post-training, we validate training strategies and algorithmic techniques in the simulated environment, then deploy verified optimal policies to the real environment for final training. 
The choice of environments plays a crucial role, agentic intelligence emerges not from a single wolrd, but from carefully chosen environments.

Agent training fundamentally depends on synthetic data and environment interaction.
Based on these design principles, we then introduce Tongyi DeepResearch in detail below.
\section{Tongyi DeepResearch}
\subsection{Formulation}

We formally define the Tongyi DeepResearch's rollout at each timestep $t$ through three fundamental components:  

\begin{itemize}[leftmargin=1.5em, itemsep=2pt, topsep=1pt, parsep=0pt]
    \item \textbf{Thought ($\tau_t$)}: The internal cognitive process of the agent. This includes analyzing the current context, recalling information from memory, planning subsequent steps, and engaging in self-reflection to adjust its strategy.
    \item \textbf{Action ($a_t$)}: An external operation executed by the agent to interact with its environment.
    Tongyi DeepResearch is equipped with a versatile set of tools that define its action space, enabling it to interact with a wide range of information sources:
    \textit{Search}, \textit{Visit}, \textit{Python Interpreter}, \textit{Google Scholar} and \textit{File Parser}.
    Actions encompass all intermediate tool calls and the final response to the user. In a given trajectory, intermediate actions ($a_t$ where $t < T$) are tool calls, while the final action, $a_T$, constitutes the generation of an in-depth report for the user.
    
    \item \textbf{Observation ($o_t$)}: The feedback received from the environment after an action is performed. This new information is used to update the agent's internal state and inform its next thought.
\end{itemize}

Based on the fundamental components above, we define two different rollout types as follows:

\paragraph{ReAct.} Tongyi DeepResearch's architecture is fundamentally based on the vanilla ReAct~\citep{yao2023react} framework, which synergizes reasoning and acting. 
In this paradigm, the agent generates both a reasoning trace (Thought) and a subsequent Action in an interleaved manner. 
This process forms a trajectory, $\mathcal{H}_T$, which is a sequence of thought-action-observation triplets:
\begin{equation}
    \mathcal{H}_T=(\tau_0,a_0,o_0,\dots, \tau_i,a_i,o_i, \dots,\tau_{T},a_{T}), 
\end{equation}
where $a_T$ represents the final answer to the given task.
At any given step $t\leq T$, the agent's policy, $\pi$, generates the current thought $\tau_t$ and action $a_t$ based on the history of all previous interactions, $\mathcal{H}_{t-1}$:
\begin{equation}
\tau_t, a_t \sim \pi(\cdot | \mathcal{H}_{t-1}).
\end{equation}
While more complex single and multi-agent paradigms have emerged, our choice of ReAct is a deliberate one, rooted in its simplicity and alignment with fundamental principles. This decision is informed by "The Bitter Lesson"~\citep{sutton2019bitter}, which posits that general methods leveraging scalable computation ultimately outperform approaches that rely on complex, human-engineered knowledge and intricate designs. Frameworks that require extensive, specialized prompt engineering or possess rigid operational structures risk becoming obsolete as the intrinsic capabilities of models scale~\citep{li2025lara}.

\paragraph{Context Management.}  
The execution of long-horizon tasks is fundamentally constrained by the finite length of the agent's context window. 
To mitigate the risk of context overflow and ensure task focus, we propose the context management paradigm \citep{qiao2025webresearcher,chen2026iterresearch}, which employs a dynamic context management mechanism based on Markovian state reconstruction.
Within this framework, the agent is not conditioned on the complete history.
Instead, at each step $t$, it is conditioned on a strategically reconstructed workspace containing only essential elements: the question $q$, an evolving report $S_t$ serving as compressed memory, and the immediate context from the last interaction ($a_t$ and $o_t$).
This Markovian structure enables the agent to maintain consistent reasoning capacity across arbitrary exploration depths while naturally circumventing the degradation.
For every step $0<t<T$, this core update process can be formalized as:
\begin{equation}
S_t, \tau_{t+1}, a_{t+1} \sim \pi(\cdot | S_{t-1}, a_t, o_t).
\end{equation}
This context management paradigm is particularly crucial, it not only prevents context suffocation but also enforces structured reasoning by requiring the agent to explicitly synthesize and prioritize information at each step. 
This design naturally aligns with human research patterns, where periodic synthesis and reflection are essential for maintaining coherent long-term investigation.

\subsection{Overall Training Recipe}
The system is initialized from the pretrained base model Qwen3-30B-A3B-Base\footnote{\url{https://huggingface.co/Qwen/Qwen3-30B-A3B-Base}}.
Tongyi DeepResearch is developed through an end-to-end training framework that integrates agentic mid-training and post-training, enabling scalable reasoning and information seeking across complex research tasks. 
This establishes a new paradigm for training agentic models.
We first present the mid-training process in Section~\ref{sec:mid-training}, followed by the post-training stage in Section~\ref{sec:post-training}.

\begin{figure}[h]
    \centering
\includegraphics[width=0.95\linewidth]{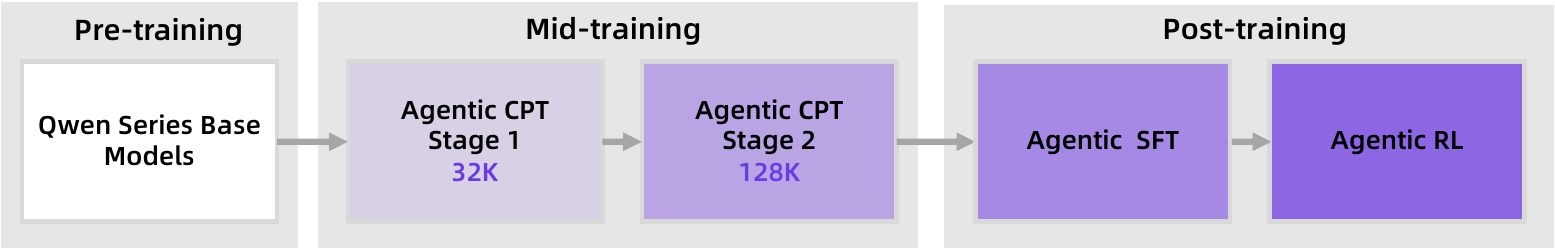}
    \caption{Training pipeline of Tongyi DeepResearch.}
    \label{fig:abs_fig}
\end{figure}

% \subsection{Agentic Continual Pre-training}
\subsection{Agentic Mid-training}
\label{sec:mid-training}
\subsubsection{Training Configuration}
Tongyi DeepResearch employs a two-stage \textbf{Agentic Continual Pre-training (Agentic CPT)}~\citep{agentfounder2025} as its core \textit{mid-training} phase. 
This phase functions as a critical bridge connecting pre-trained models and agentic post-training. 
Its primary objective is to provide a base model endowed with a strong inductive bias for agentic behavior, while simultaneously preserving broad linguistic competence. 
To achieve this, the optimization process is driven by the standard \textit{Next-Token Prediction} loss function.

The design of this phase is strategically optimized for both efficiency and progressive capability scaling.
We initiate with a \textbf{32K} context length in the first stage, before expanding to \textbf{128K} in the second.
This expanded context window is specifically leveraged in the second stage, where we introduce a substantial corpus of long-sequence (64K-128K) agentic behavior data.
This approach is critical for enhancing the model's capacity for coherent, long-horizon reasoning and action. 
Throughout both stages, a small proportion of general pre-training data is interleaved, ensuring the model acquires specialized agentic competence without sacrificing its foundational generalization capabilities.

\subsubsection{Large-scale Agent Behavior Data Synthesis}

\begin{figure}[h]
    \centering
\includegraphics[width=0.95\linewidth]{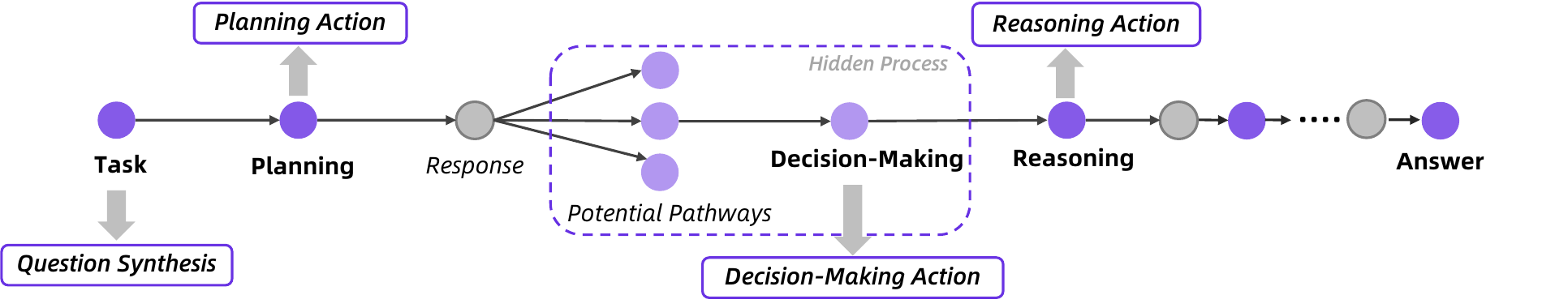}
    \caption{Large-scale agent behavior data synthesis for agentic continual pre-training.}
    \label{fig:large_scale_cpt}
\end{figure}

In Agentic CPT, we synthesize data across the complete lifecycle of agent workflows as shown in Figure~\ref{fig:large_scale_cpt}.
A typical agent workflow begins with a problem, iteratively cycles through reflection and action, and ultimately converges on a final solution. 
To comprehensively capture this process, we synthesize data for the critical steps that constitute the agent's operational cycle: Question Synthesis, Planning Action, Reasoning Action, and Decision-Making Action. 
Note that while decision-making is often implicit within agent cycles, we explicitly model it as a distinct action type in our synthesis framework.

\textbf{Large-scale Multi-style Question Synthesis}. 
Grounded in continuously updated open-world knowledge, we construct an entity-anchored open-world memory. This memory consolidates diverse real-world knowledge sources, such as web-crawled data and agent interaction trajectories, into structured representations of entities and their associated knowledge. Building upon this foundation, we sample entities along with their related knowledge to generate diverse questions that embed specific behavioral pattern requirements, such as multi-hop reasoning questions and numerical computation questions.

\textbf{Planning Action}. Planning refers to problem decomposition and first-step action prediction. A key insight is that planning accuracy is highly correlated with whether an agent can successfully complete a task. Thus, we employ open-source models to analyze, decompose, and predict initial actions for the synthesized questions. Furthermore, we leverage the entities and associated knowledge used in question construction as the basis for rejection sampling, thereby ensuring high-quality planning outputs.

\textbf{Reasoning Action}. Logical reasoning and knowledge integration over heterogeneous data is foundational for agents solving complex tasks. When external tools return massive unstructured responses, whether models can distill critical knowledge from noise and construct coherent reasoning paths directly determines task outcomes. To this end, given a question and its dependent knowledge, we guide large models through a two-stage process to generate complete reasoning chains, with a dual filtering mechanism based on reasoning length and answer consistency to ensure quality.

\textbf{Decision-Making Action}. Each step of an agent's thinking and action is essentially an implicit decision-making process. Specifically, each decision point encompasses multiple potential reasoning and action paths, from which the agent must select the most promising solution. To capture this critical mechanism, we explicitly model this decision-making process. First, based on existing demonstration trajectories, we thoroughly explore the feasible action space at each step. Second, we reconstruct the original trajectories into multi-step decision sequences while preserving the original decision choices.

% \begin{figure}[h]
%     \centering
%     \includegraphics[width=0.95\linewidth]{pics/decision-making-action.pdf}
%     \caption{Synthesis Pipeline of Deceision Making Action Data.}
%     \label{fig:abs_fig}
% \end{figure}

\paragraph{General Function-calling Data Synthesis via Environment Scaling.}
To enhance our model’s general agentic capability, we systematically scale the function-calling data through environment scaling.
The breadth of function-calling competence is closely tied to the diversity of environments in which agents are trained~\citep{fang2025towards}. 
We also scale up environments as a step towards advancing general agentic intelligence.
In designing environment construction and scaling, we follow the principle that the core of an agent lies in its capacity for environment interaction, with each environment instantiated as a \textit{read}–\textit{write} database.
We design a scalable framework that automatically constructs heterogeneous environments that are fully simulated, systematically broadening the space of function-calling scenarios.
The produced data are incorporated into the model’s mid-training phase.

\subsection{Agentic Post-training}
\label{sec:post-training}
The post-training pipeline comprises three stages: data synthesis, supervised fine-tuning for cold start, and agentic reinforcement learning.

\subsubsection{High-quality Data Synthesis}
\begin{figure}[h]
    \centering
    \includegraphics[width=0.75\linewidth]{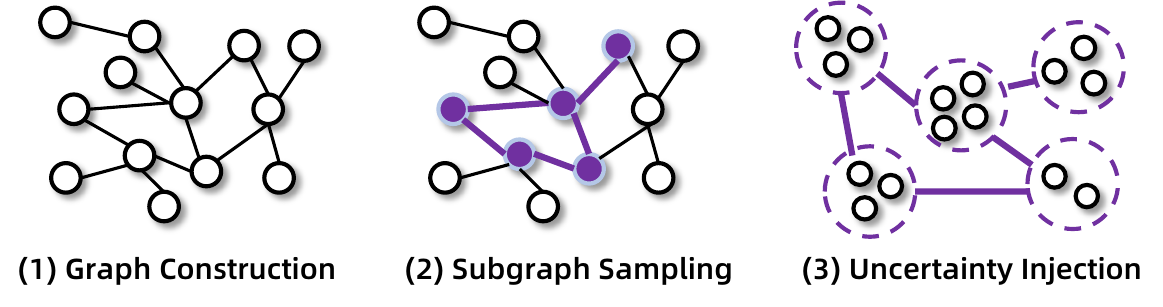}
    \caption{High-quality data synthesis pipeline.}
    \label{fig:data}
\end{figure}
We develop an end‑to‑end solution for synthetic data generation to generate complex, high‑uncertainty and super-human level question and answer pairs~\citep{li2025websailor,li2025websailorv2}, as shown in Figure~\ref{fig:data}. 
This fully automated process requires no human intervention to construct super‑human quality datasets, designed to push the boundaries of agent performance. 
The process begins by constructing a highly interconnected knowledge graph via random walks, leveraging web search to acquire relevant knowledge, and isomorphic tables from real‑world websites, ensuring a realistic information structure.
We then sample subgraphs and subtables to generate initial questions and answers. 
The pivotal step involves strategically increasing the uncertainty within the question to enhance its difficulty~\citep{wu2025webdancer}.
This practical approach is grounded in a complete theoretical framework, where we formally model QA difficulty as a series of controllable "atomic operations" (\textit{e.g.}, merging entities with similar attributes) on entity relationships, allowing us to systematically increase complexity.
To further reduce inconsistencies between the organized information structure and the reasoning structure of QA, enable more controllable difficulty and structure scaling of reasoning, we proposed a formal modeling of the information‑seeking problem based on set theory~\citep{tao2025webshaper}.
With this formalization, we develop agents that expands the problem in a controlled manner, and minimizes reasoning shortcuts and structural redundancy, leading to further improved QA quality. 
Moreover, this formal modeling also allows for efficient verification of QA correctness, effectively addressing the challenge of validating synthetic information‑seeking data for post‑training.

We also develop an automated data engine to scale the generation of PhD-level research questions~\citep{qiao2025webresearcher}.
Starting from a multi-disciplinary knowledge base, it creates seed QA pairs requiring multi-source reasoning.
These seeds undergo iterative complexity upgrades, where a question-crafting agent, equipped with the corresponding tool, progressively expands scope and abstraction.
Each iteration refines and compounds prior outputs, enabling a systematic and controllable escalation of task difficulty.

\subsubsection{Supervised Fine-tuning for Cold Start}
The initial phase of our agentic post-training pipeline is a supervised fine-tuning (SFT) stage, designed to equip the base model with a robust initial policy prior to reinforcement learning. 
Starting from our synthesized high-quality QA data, we obtain training trajectories that cover the complete thought process and tool responses generated by high-performing open-source models, which are then subjected to a rigorous rejection sampling protocol.
This comprehensive filtering process guarantees that only high-quality trajectories exhibiting diverse problem-solving patterns are retained.

\paragraph{Mixed Training Paradigm.} 
The cold stage training leverages data from two different formulations to enhance model robustness and generalization.
For the React Mode, the training samples take the historical state $\mathcal{H}_{t_1}$ as input, and output the corresponding thought $\tau_i$ and tool call $a_i$ for the current step.
For our Context Management Mode , the training samples take as input the previous step’s trajectory summary $S_{t-1}$, tool call $a_{i-1}$, and tool response $o_{i-1}$, and output the current step’s trajectory summary, thought $\tau_i$, and tool call $a_i$.
The Context Management Mode data particularly strengthens the agent's capabilities in state analysis and strategic decision-making, as it requires the model to synthesize complex observations into coherent summaries while maintaining task focus across extended trajectories.
This synthesis-oriented training enables more deliberate reasoning patterns compared to purely ReAct.
We adopt a two-stage training strategy based on context length.
In the first stage, the context length is set to \textbf{40K}, and the training data consist of ReAct Mode samples with context lengths shorter than 40K, along with all Context Management Mode samples (as they are all within 40k).
In the second stage, the context length is extended to \textbf{128K}, and the training data include ReAct Mode samples with context lengths between 40K and 128K, as well as a small portion of 40K data for stability.

\subsubsection{Agentic Reinforcement Learning}
\label{sec:rl}
\begin{figure}[h]
    \centering
    \includegraphics[width=0.95\linewidth]{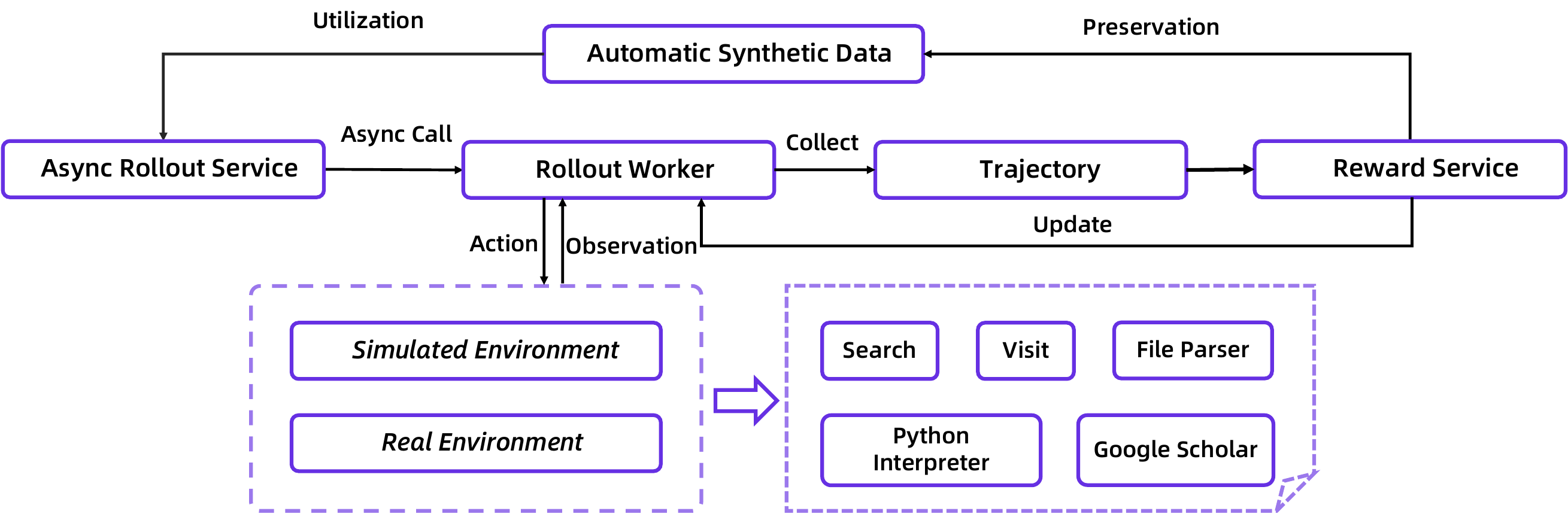}
    \caption{An overview of our agentic reinforcement learning framework.}
    \label{fig:agentic_rl}
\end{figure}

To advance the model's capabilities toward more robust and reliable planning and searching in a complex web environment, we apply an agentic RL framework, which is illustrated in Figure~\ref{fig:agentic_rl}.
In this framework, the model generates a complete task attempt (a "rollout") and receives a reward if its final answer matches the ground truth (RLVR)~\citep{r1}.
Throughout this agentic RL procedure, the model continuously interacts with the environment (simulated or real-world), iteratively refining its policy with each iteration, and, in turn, using that improved policy to curate a new, higher-quality set of training data.
\paragraph{Real-world Environment.} 
Our agent's toolkit is a complex system that integrates several specialized tools\footnote{The details for each tool are shown in Appendix~\ref{app:env}.}: 
(1) \textbf{Search}, 
(2) \textbf{Visit}, 
(3) \textbf{Python Interpreter},
(4) \textbf{Google Scholar},
(5) \textbf{File Parser}.
The end-to-end reliability of this system is paramount. 
The inherent volatility of external APIs, encompassing high latency, outright failures, and inconsistent returns, threatens to corrupt our training trajectories. 
This data contamination makes it nearly impossible to diagnose performance issues, obscuring whether a poor outcome is caused by a weakness in the agent's policy or by the instability of the environment itself.
To ensure reliable tool use during agent training and evaluation, we developed a unified sandbox.
This interface is built around a central scheduling and management layer that orchestrates every tool call. 
For each tool, we implement robust concurrency controls and fault-tolerance mechanisms, such as proactive QPS rate constraints, result caching, automatic timeout-and-retry protocols, graceful service degradation for non-critical failures, and seamless failover to backup data sources (\textit{e.g.}, a backup search API). This design abstracts the tool invocation into a deterministic and stable interface for the agent and thereby insulates the training loop from real-world stochasticity while also significantly reducing operational costs.
This design abstracts tool invocation into a deterministic interface, providing a stable and fast experience that is crucial for preventing tool errors from corrupting the agent's learning trajectory.

\paragraph{Simulated Environment.} 
Directly utilizing real-world web environment APIs presents numerous practical problems\footnote{Queries per second (QPS) impact significantly degrade our development efficiency and compromise the reliability during our early-stage ablation studies.}.
We first build an offline environment based on the 2024 Wikipedia database and develop a suite of local RAG tools to simulate the web environment. 
We then reuse the data synthesis pipeline to create a high-quality, structurally complex QA specifically for this offline environment.
This provides us with a low-cost, high-efficiency, and fully controllable platform that enables high-frequency, rapid experimentation, thereby greatly accelerating our development and iteration process.

\paragraph{On-Policy Asynchronous Rollout Framework.} 
The iterative nature of agentic rollouts, which require numerous interactions with the environment, creates a significant bottleneck that slows down the entire RL training process. 
To overcome this, we implement a custom, step-level asynchronous RL training loop built on the rLLM framework~\citep{rllm2025}. 
Our solution utilizes two separate asynchronous online servers, with one for model inference and another for tool invocation.
A centralized interaction handler then processes the outputs from both, formatting the feedback into a unified message list. 
This architecture allows multiple agent instances to interact with the environment in parallel, each completing its rollout independently.

\paragraph{RL Training Algorithm.} 
Our RL algorithm is a tailored adaptation of GRPO~\citep{shao2024deepseekmath}:
\begin{equation}
\begin{aligned}
\mathcal{J}(\theta) =\quad& \mathbb{E}_{(q,y)\sim \mathcal{D}, \{\mathcal{H}^i\}_{i=1}^G\sim \pi_{\theta_\text{old}}(\cdot\mid context)}\\&
\Bigg[\frac{1}{\sum_{i=1}^{G}|\mathcal{H}^i|}\sum_{i=1}^{G}\sum_{j=1}^{|\mathcal{H}^i|} 
\min \Big( r_{i,j}(\theta) \hat{A}_{i,j},  
\ \text{clip} \Big( r_{i,j}(\theta), 1 - {\varepsilon_{low}}, 1 + {\varepsilon_{high}} \Big) \hat{A}_{i,j} \Big) \Bigg], 
\label{eq:dapoloss}
\end{aligned}
\end{equation}
where $(q,y)$ is the question-answer pair, $r_{i,j}(\theta)$ is the importance sampling ratio (remains 1.0 for strictly on-policy training), and $\hat{A}_{i,j}$ is an estimator of the advantage at token $j$:
\begin{equation}
    r_{i,j}(\theta)=\frac{\pi_{\theta}(\mathcal{H}^{i,j} \mid context)}{\pi_{\theta_{\text{old}}}(\mathcal{H}^{i,j} \mid context)},\quad\hat{A}_{i,j} = R_i - \text{mean}(\{R_i\}_{i=1}^G).
\label{eq:advantage_calculation}
\end{equation}

We employ a strict on-policy regimen, where trajectories are consistently sampled using the most up-to-date policy, ensuring that the learning signal is always relevant to the model’s current capabilities. 
The reward is a pure 0 or 1 signal of answer correctness. We \textbf{do not} include a format reward (\textit{e.g.}, 0.1 for format correctness) because the preceding cold start stage ensures the model is already familiar with the required output format. 
Following DAPO~\citep{yu2025dapo}, we apply the token-level policy gradient loss in the training objective and clip-higher strategy to encourage more exploration.
To further reduce variance in the advantage
estimation, we adopt a leave-one-out strategy~\citep{chen2025reinforcement}.
Furthermore, we observed in preliminary experiments that directly optimizing on an unfiltered set of negative rollouts significantly degrade training stability and can lead to policy collapse after extended training. To mitigate this, we selectively exclude certain negative samples from the loss calculation, for instance,
those that do not yield a final answer because they exceed a length limit. The primary motivation for these modifications is not algorithmic novelty but the pragmatic pursuit of a more efficient and stable training paradigm.

\paragraph{Automatic Data Curation.}
We optimize data in real time, guided by training dynamics to generalize to out‑of‑distribution scenarios through self‑exploration. 
This optimization is achieved through a fully automated data filtering pipeline that dynamically adjusts the training set based on the improved policy model. 
Specifically, our process begins with a large dataset, $\mathcal{D}$. 
We use the initial SFT model as a baseline policy to sample multiple solution attempts, or rollouts, for each problem.
We then create an initial training set, $\mathcal{D'}$, by filtering out problems where the model either always fails or always succeeds, as these will offer no learning signal for RL training. This leaves us with a focused subset of problems of moderate difficulty.
During RL training, we continuously monitor the problems in $\mathcal{D'}$ by their latest rollouts to see if they have become too easy for the improved policy model.
In parallel, a separate process uses intermediate checkpoints of the policy model to sample from the entire original dataset, $\mathcal{D}$. 
This background process identifies and collects a backup pool of new problems that have become moderately difficult for the now-stronger model. 
When the training reaches a certain step count or the reward plateaus, we refresh the active training set $\mathcal{D'}$ by removing the mastered problems and incorporating new, challenging ones from the backup pool. The entire data filtering and refreshment pipeline runs independently, never interrupting the main RL training loop. This design allows us to automatically evolve both the policy model and its training data, ensuring consistently high training efficiency and stability.

Through our experiments, we arrive at a critical insight: \textbf{the success of agentic RL depends more on the quality of the data and the stability of the training environment than on the specific algorithm being used}. Consequently, we concentrate our efforts on designing a stable environment and curating high-quality data, making only a few essential modifications to the algorithm itself, mainly for the purpose of stabilizing the training process.

\subsubsection{Model Merging}
We employ model merging at the last stage of the pipeline.
This approach is built on the key insight that when different model variants are derived from the same pre-trained model, their parameters can be effectively combined through averaging or interpolation~\citep{wang2025ui}.
Specifically, our process involves selecting several model variants that originate from the same base model but exhibit different capability preferences.
We then create the final merged model by computing a weighted average of their parameters:
\begin{equation}
\theta_{\mathrm{merged}} = \sum_{k} \alpha_k \cdot \theta^{(k)}, \quad \text{s.t.} \quad \sum_{k} \alpha_k = 1, \ \alpha_k \ge 0.
\end{equation}
where $\theta^{(k)}$ represents the parameters of the $k$-th model variant, and $\alpha_k$ is its corresponding merge weight.
Empirically, this interpolation strategy not only preserves the core strengths of each contributing model but also equips the merged model with robust generalization abilities. 
In complex scenarios requiring a synthesis of these varied capabilities, the merged model performs comparably to the best-performing source model in its respective area of strength, all without incurring additional optimization costs.
\section{Experiments}
\subsection{Experimental Setup}
\noindent \textbf{Backbones.}
We evaluate Tongyi DeepResearch on seven public information-seeking benchmarks spanning long-term reasoning and long-horizon tool use.
The model is compared against two families of systems: 1) LLM-based ReAct agents: GLM-4.5~\citep{zeng2025glm}, Kimi-K2~\citep{team2025kimi}, DeepSeek-V3.1~\citep{deepseekv3.1}, Claude-4-Sonnet~\citep{claude4}, OpenAI o3/o4-mini~\citep{o3}) and 2) end-to-end deep-research agents: OpenAI DeepResearch~\citep{dr}, Gemini DeepResearch~\citep{geminiresearch}, Kimi Researcher~\citep{kimiresearcher}.

\noindent \textbf{Benchmarks.}
We follow each benchmark’s official evaluation protocol. 
The benchmarks cover: (1) Humanity’s Last Exam~\citep{phan2025humanity}; (2) BrowseComp~\citep{bc_en} and BrowseComp-ZH~\citep{bc_zh}; (3) GAIA~\citep{mialon2023gaia}; (4) xBench-DeepSearch~\citep{xbench}; (5) WebWalkerQA~\citep{wu2025webwalker}; (6) FRAMES~\citep{krishna2025fact}; and (7) xbench-DeepSearch-2510.

All scores are computed with the official scripts released by each benchmark. The details of evaluation are presented in Appendix~\ref{app:eval}.

\textbf{Evaluation.} We adopt fixed inference parameters to ensure stability and reproducibility across evaluations: temperature = 0.85, repetition penalty = 1.1, and top-p = 0.95.
A maximum of 128 tool invocations is allowed per task, and the context length is constrained to 128K tokens.
Each benchmark is evaluated three times independently, and we report the average performance (\texttt{Avg}@3) as the main metric.
For completeness, we also report the best \texttt{Pass}@1 (best result over 3 runs) and \texttt{Pass}@3 results in the subsequent analysis.
All results are obtained on September 16, 2025, except for xbench-DeepSearch-2510, which is evaluated on October 28, 2025.

\textbf{Reproduce.} Tongyi DeepResearch operates utilizing an action space that includes the Search, Visit, Python, Scholar, and File Parser tools.
We release official reproduction scripts on GitHub\footnote{\url{https://github.com/Alibaba-NLP/DeepResearch}}, along with the complete tool implementations and prompt configurations.

\subsection{Main Results}
\begin{table}[htbp]
\centering
\caption{Performance comparison on various benchmarks.}
\label{tab:main_results}
\resizebox{\textwidth}{!}{
\begin{tabular}{lccccccc}
\toprule
\textbf{Benchmarks} & \textbf{Humanity's} & \textbf{Browse} & \textbf{Browse} & \textbf{GAIA} & \textbf{xbench} & \textbf{WebWalker} & \textbf{FRAMES}  \\
& \textbf{Last Exam} & \textbf{Comp} & \textbf{Comp-ZH} & & \textbf{DeepSearch} & \textbf{QA} & \\
\midrule
\multicolumn{8}{l}{\textit{LLM-based ReAct Agent}} \\
\midrule
GLM 4.5 & 21.2 & 26.4 & 37.5 & 66.0 & 70.0 & 65.6 & 78.9 \\
Kimi K2 & 18.1 & 14.1 & 28.8 & 57.7 & 50.0 & 63.0 & 72.0 \\
DeepSeek-V3.1 & 29.8 & 30.0 & 49.2 & 63.1 & 71.0 & 61.2 & 83.7 \\
Claude-4-Sonnet & 20.3 & 12.2 & 29.1 & 68.3 & 65.0 & 61.7 & 80.7 \\
OpenAI o3 & 24.9 & 49.7 & 58.1 & -- & 67.0 & 71.7 & 84.0 \\
OpenAI o4-mini & 17.7 & 28.3 & -- & 60.0 & -- & -- & -- \\
\midrule
\multicolumn{8}{l}{\textit{DeepResearch Agent}} \\
\midrule
OpenAI DeepResearch & 26.6 & 51.5 & 42.9 & 67.4 & -- & -- & -- \\
Gemini DeepResearch & 26.9 & -- & -- & -- & -- & -- & -- \\
Kimi Researcher & 26.9 & -- & -- & -- & 69.0 & -- & 78.8 \\
\midrule
\rowcolor{blue!10} Tongyi DeepResearch (30B-A3B) & \textbf{32.9} & \textbf{43.4} & \textbf{46.7} & \textbf{70.9} & \textbf{75.0} & \textbf{72.2} & \textbf{90.6}\\
\bottomrule
\end{tabular}
}
\end{table}

Table~\ref{tab:main_results} presents the performance of Tongyi DeepResearch compared with a broad range of state-of-the-art LLM-based agents and proprietary deep research systems across multiple benchmarks, including Humanity’s Last Exam, BrowseComp, BrowseComp-ZH, GAIA, xbench DeepSearch, WebWalker QA, and FRAMES.
Tongyi DeepResearch achieves the highest scores on nearly all evaluated benchmarks, demonstrating strong generalization across both English and Chinese tasks. 
It consistently surpasses both open and closed commercial systems, including OpenAI o3, DeepSeek-V3.1, and Gemini DeepResearch.
On the newly released xbench-DeepSearch-2510, Tongyi DeepResearch ranks just below ChatGPT-5-Pro, demonstrating competitive performance at the forefront of the field.
Notably, these gains are achieved with only 3.3 billion activated parameters per token, underscoring the model’s efficiency and scalability.
In aggregate, Tongyi DeepResearch sets a new state of the art among open-source deep research agents, narrowing and in some cases even surpassing the performance of frontier proprietary systems while maintaining superior interpretability and computational efficiency.

\subsection{Heavy Mode}
\begin{figure}[h]
    \centering
    \includegraphics[width=\linewidth]{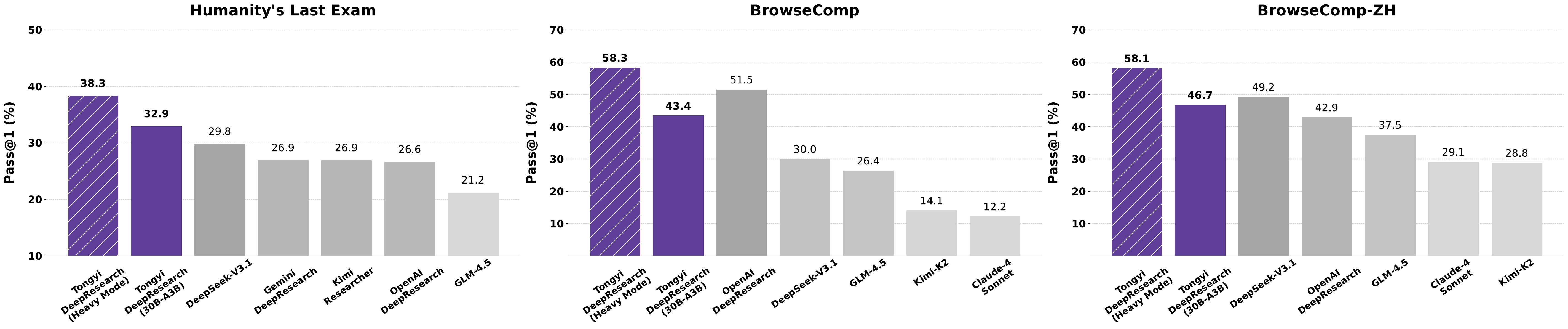}
    \caption{Performance comparison between Tongyi DeepResearch Heavy Mode and state-of-the-art models. 
    }
    \label{fig:heavy_mode}
\end{figure}

To further unlock the potential of deep research agents, we introduce the \textbf{Heavy Mode}, which leverages test-time scaling through a Research-Synthesis framework built upon the context management paradigm~\citep{chen2026iterresearch}. Given that DeepResearch involves multi-round tool calls and intensive reasoning, directly aggregating contexts from multiple trajectories is computationally prohibitive. Our Heavy Mode addresses this challenge through strategic parallelization and synthesis.

\paragraph{Parallel Research Phase.} 
We deploy $n$ parallel agents, each following the context management paradigm but exploring diverse solution paths through different tool usage and reasoning strategies. 
Each agent $u$ independently processes the question $q$ and produces a final report and answer:
\begin{equation}
(S^u_T, \text{answer}_u) = \text{Agent}_u(q), \quad u \in [1,n]
\end{equation}
where $S^u_T$ represents the final report summary from agent $u$ after $T$ iterations, encapsulating the complete reasoning trajectory in compressed form.

\paragraph{Integrative Synthesis Phase.}
A synthesis model consolidates all parallel findings to produce the final answer:
\begin{equation}
\text{answer}_{\text{final}} = \text{Synthesis}\left(\{(S^u_T, \text{answer}_u)\}_{u=1}^n\right),
\end{equation}

The key advantage of this approach lies in the compressed nature of context management reports $S^u_T$.
Unlike traditional methods that would require aggregating full trajectories (potentially exceeding context limits with just 2-3 agents), our approach enables the synthesis model to assess $n$ diverse solution strategies within a manageable context window. Each report $S^u_T$ preserves the essential reasoning logic and findings while discarding redundant intermediate steps, enabling effective test-time scaling.

As shown in Figure~\ref{fig:heavy_mode}, our Heavy Mode achieves state-of-the-art performance on Humanity's Last Exam (38.3\%) and BrowseComp-ZH (58.1\%), while remaining highly competitive on BrowseComp (58.3\%). These substantial improvements validate the effectiveness of our heavy mode based on context management in leveraging test-time compute through parallel exploration and intelligent aggregation.

\subsection{Detailed Analysis}

\begin{figure}[h]
    \centering
    \includegraphics[width=0.65\linewidth]{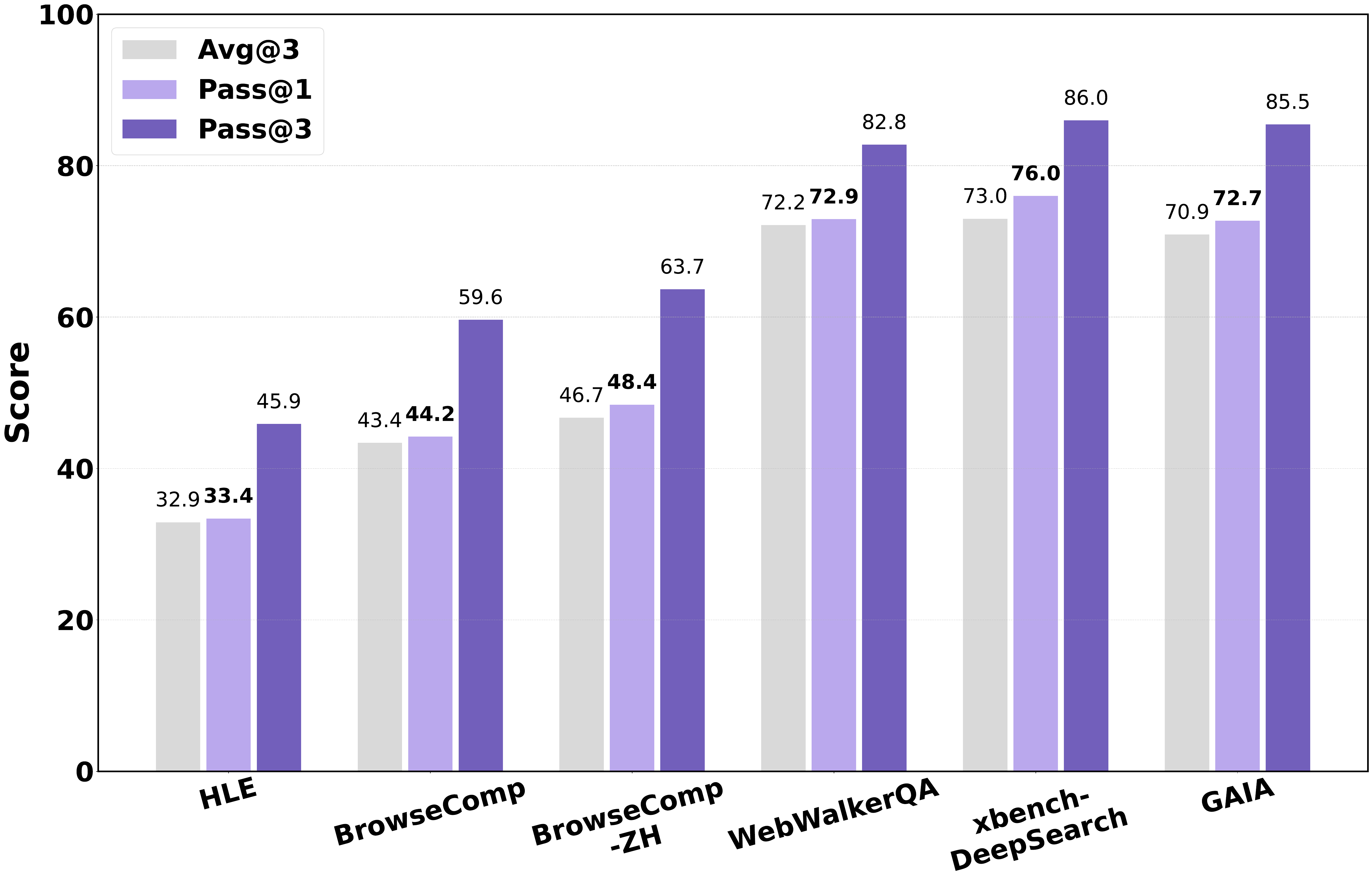}
    \caption{Detailed evaluation results using \texttt{Avg}@3, \texttt{Pass}@1 and \texttt{Pass}@3 metric.}
    \label{fig:detailed}
\end{figure}

\textbf{Pass@1 and Pass@3 Performance.} We report the \texttt{Avg}@3 performance in Table~\ref{tab:main_results}.
Given the dynamic and complex nature of agent environments, we further conduct a fine-grained analysis of \texttt{Pass}@1 (over three runs) and \texttt{Pass}@3 in Figure~\ref{fig:detailed}. 
Despite the unstable evaluation environment, our final \texttt{Avg}@3 results are consistent with the \texttt{Pass}@1 (best result over 3 runs) results, demonstrating the robustness of our deep research approach. 
Our \texttt{Pass}@3 performance demonstrates the strong potential of our agent.
In particular, it achieves \textbf{59.64 on BrowseComp}, \textbf{63.67 on BrowseComp-ZH}, and \textbf{45.9 on Humanity’s Last Exam}.

\begin{figure}[h]
    \centering
    \captionsetup{skip=4pt} % 主标题与图之间距离，可调 0–6pt

    \begin{minipage}[b]{0.45\textwidth}
        \includegraphics[width=\textwidth]{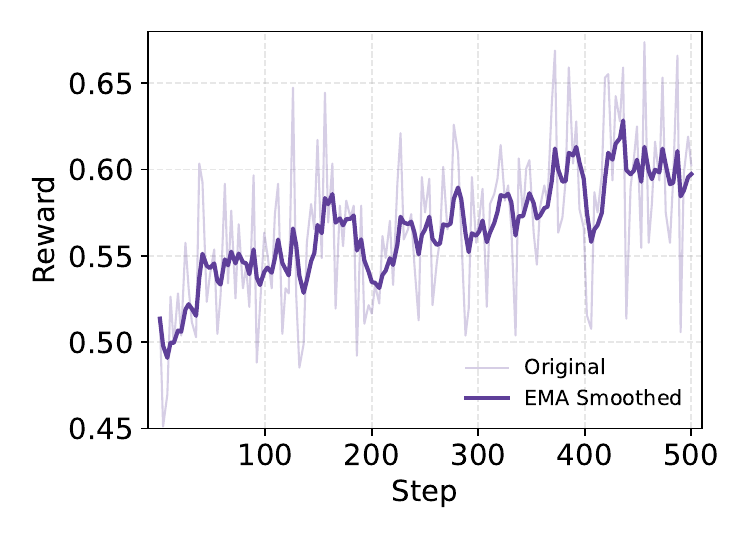}
    \end{minipage}
    % \hfill
    \begin{minipage}[b]{0.45\textwidth}
        \includegraphics[width=\textwidth]{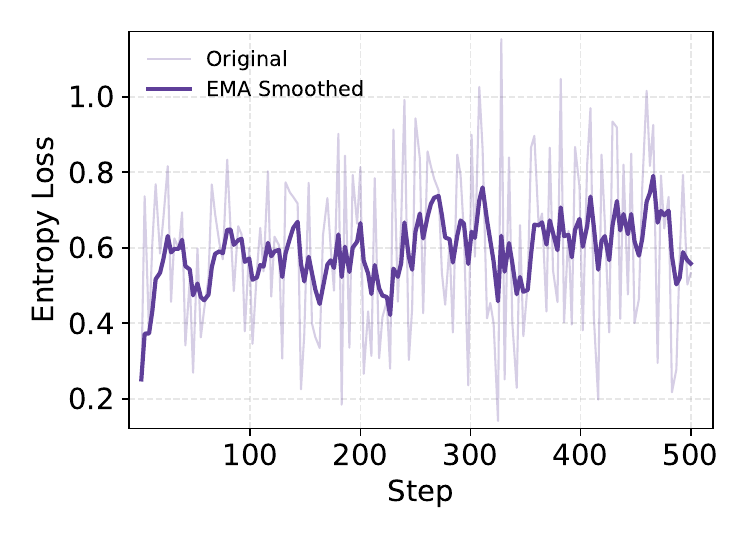}
    \end{minipage}

    \caption{Reward and entropy loss of agentic RL training.}
    \label{fig:final_rl_plots}
\end{figure}

\textbf{Training Rewards and Entropy.}
As shown in Figure~\ref{fig:final_rl_plots}, the agent's performance exhibits a clear and significant upward trend with training, confirming effective policy learning. The sustained nature of this improvement underscores the success of our dynamic data curation, which prevents learning from stagnating by consistently providing challenging material.
Concurrently, the policy entropy exhibits exceptional stability, converging to a consistent value after a brief initial increase and thereby avoiding both collapse and explosion. This outcome serves as strong evidence for our methodological contributions in environment design and algorithm modification, which together create the necessary conditions for a remarkably stable and effective RL training paradigm.

\textbf{Context Length of RL.}
In Figure~\ref{fig:context_len_comparison}, we analyze the impact of the model's context length on the agentic RL training process, comparing models with 32k, 48k, and 64k context limits. It is important to note that the dynamic data curation for all three experimental variants was performed using the same model with a 64k context. 
Focusing first on the reward dynamics in the left panel, we observe that all three models demonstrate effective and stable policy learning, evidenced by a monotonically increasing reward. This confirms the robustness of our training framework. However, their performance ceilings diverge significantly, which is an expected consequence of our data curation method. Because the curriculum is populated with problems deemed moderately difficult by the highly capable 64k context model, many of these problems inherently require long and complex reasoning to solve. Consequently, a clear hierarchy emerges: the 64k model, perfectly matched to its own data, achieves the highest reward. The 48k and 32k models, being increasingly constrained, are unable to solve the most complex problems in the curriculum, thus capping their maximum potential reward.

The training dynamics in the right panel reveal a more interesting story. The model with a 64k context exhibits a steady increase in average response length, learning to leverage its expansive context to build more elaborate solutions. In contrast, the model with a 48k context maintains a consistent equilibrium, improving its policy within a stable complexity budget. Most surprisingly, the model with a 32k context displays a clear downward trend in response length. This observation provides a key insight: for a model with a limited context, RL training on a curriculum designed for a more capable model can force it to discover more efficient solutions. This effect arises because our dynamic data curriculum is continuously updated using the 64k context model, a process that populates the training set with problems whose optimal solutions can be longer than 32k tokens. For the model with a 32k context, attempting these problems is likely to yield a zero-reward signal. This creates a powerful implicit incentive to discover more concise, potent action sequences that fit within its limit, thus becoming more efficient over time.

\begin{figure}[h]
    \centering
    % 全局/局部调主 caption 距离
    \captionsetup{skip=4pt} % 可改成 2pt, 0pt

    \begin{minipage}[b]{0.45\textwidth}
        \includegraphics[width=\textwidth]{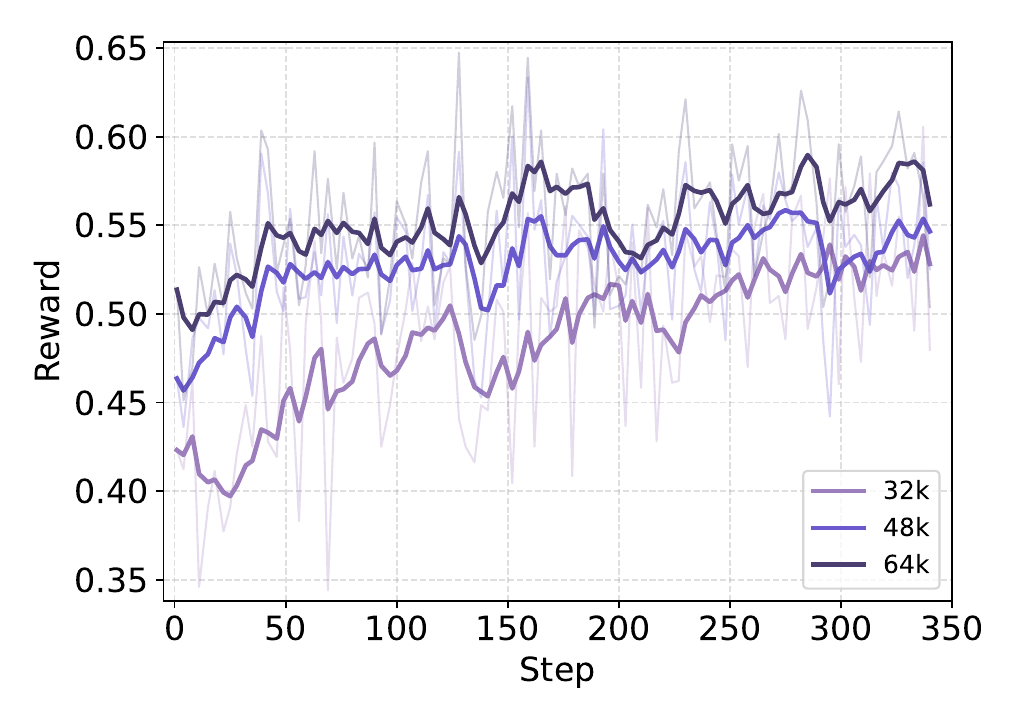}
    \end{minipage}
    % \hfill
    \begin{minipage}[b]{0.45\textwidth}
        \includegraphics[width=\textwidth]{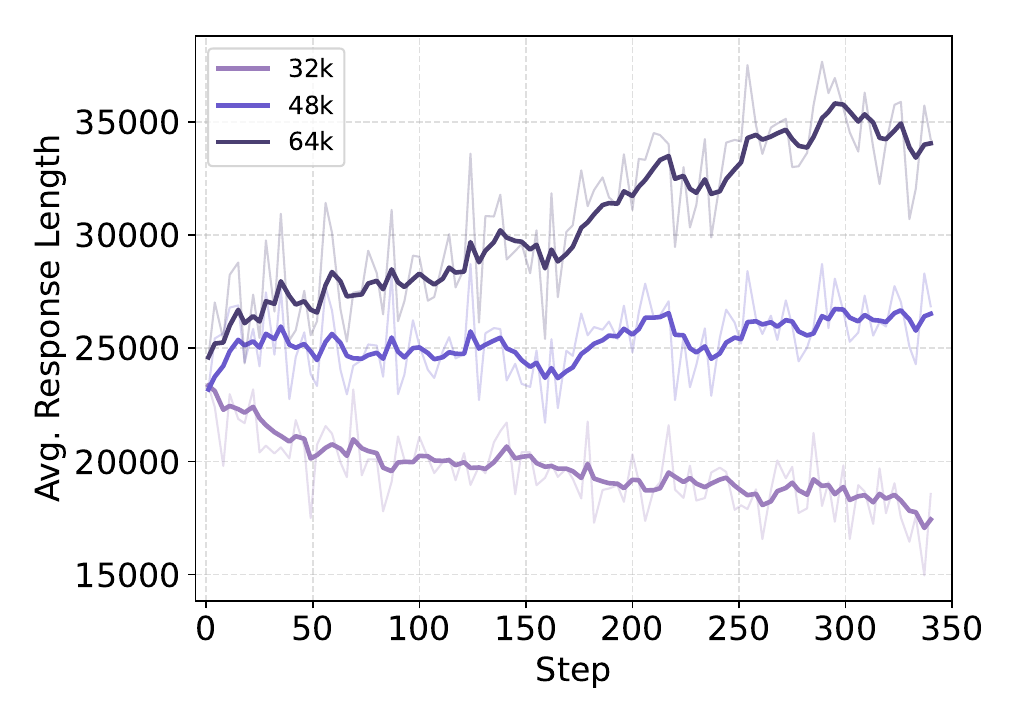}
    \end{minipage}

    \caption{Comparison of different context length limits for RL training.}
    \label{fig:context_len_comparison}
\end{figure}

\begin{figure}[t] 
\centering 
\begin{subfigure}[b]{0.45\textwidth} \includegraphics[width=\textwidth]{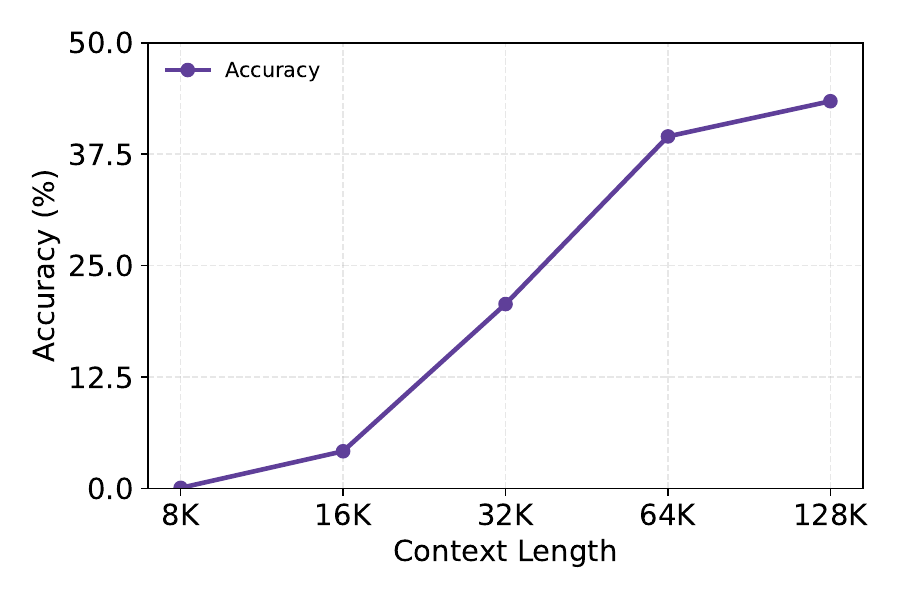} 
\caption{Interaction turns scaling for BrowseComp.}
\label{fig:scale} 
\end{subfigure} 
\begin{subfigure}[b]{0.45\textwidth} \includegraphics[width=\textwidth]{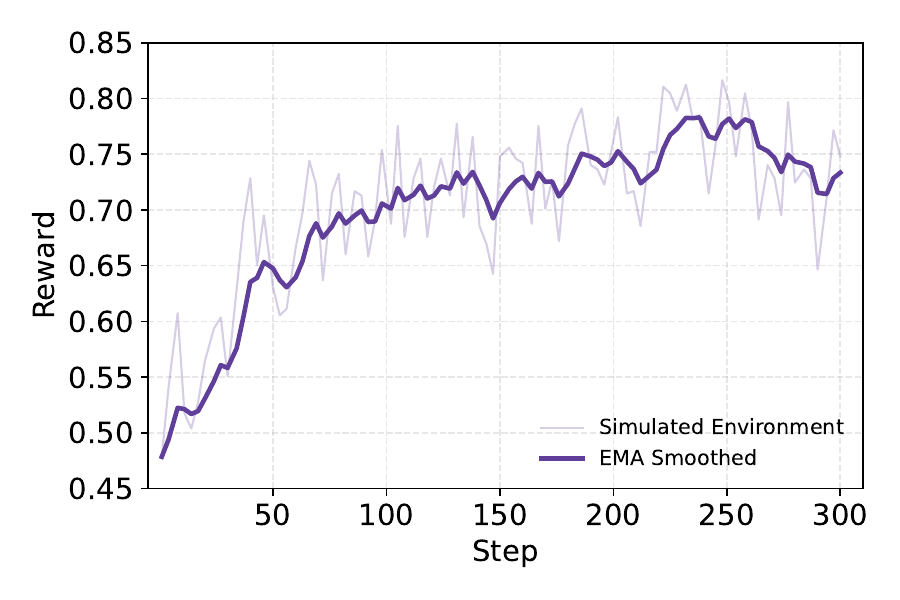} 
\caption{Reward in the simulated environment.}
\label{fig:simulated} 
\end{subfigure} 
\caption{Detailed analysis on interaction scaling and simulated environments.} \label{fig:context_len_comparison} 
\vspace{-5mm}
\end{figure}

\textbf{Interaction Test-time Scaling.} 
Unlike conventional models, the DeepResearch agent primarily relies on interactions with the environment to acquire information and accomplish tasks. 
Therefore, the number of interaction turns with the environment is crucial.
While reasoning models can be scaled by increasing the number of output tokens, our approach scales along a different dimension, the number of environment interactions.
Naturally, as the number of interactions increases, the agent obtains more observations from environment, resulting in a longer context. 
Figure~\ref{fig:scale} illustrates our scaling curve: as the context length and number of interactions grow, the model’s performance on the BrowseComp dataset improves consistently.

\textbf{Super-human Level Synthetic Data.}
To validate the effectiveness of our synthetic data, we conducted a statistical analysis of the SFT dataset. 
Over \textbf{20\%} of the samples exceed 32k tokens and involve more than 10 tool invocations.
This demonstrates the high complexity and richness of our synthetic data.
Such high-quality, cold-start data provides the model with a strong foundation for deep reasoning and research capabilities, serving as an excellent initialization for the RL phase.
During reinforcement learning, we leverage automated data curation to make more effective use of the synthetic data.

\textbf{From Simulation to Reality.}
To rapidly validate our algorithm, we built a simulated Wiki environment that mirrors real-world conditions.
We test our adapted GRPO algorithm in this environment, and the resulting reward curve, shown in Figure~\ref{fig:simulated}, closely matches the one observed in the real environment, as shown in Figure~\ref{fig:final_rl_plots}.
This Wiki simulation environment provides functionality analogous to a "wind tunnel laboratory", enabling fast algorithm iteration and significantly improved our development efficiency.

\begin{wrapfigure}{r}{7.5cm}
\vspace{-5mm}
\centering
\includegraphics[width=0.48\columnwidth]{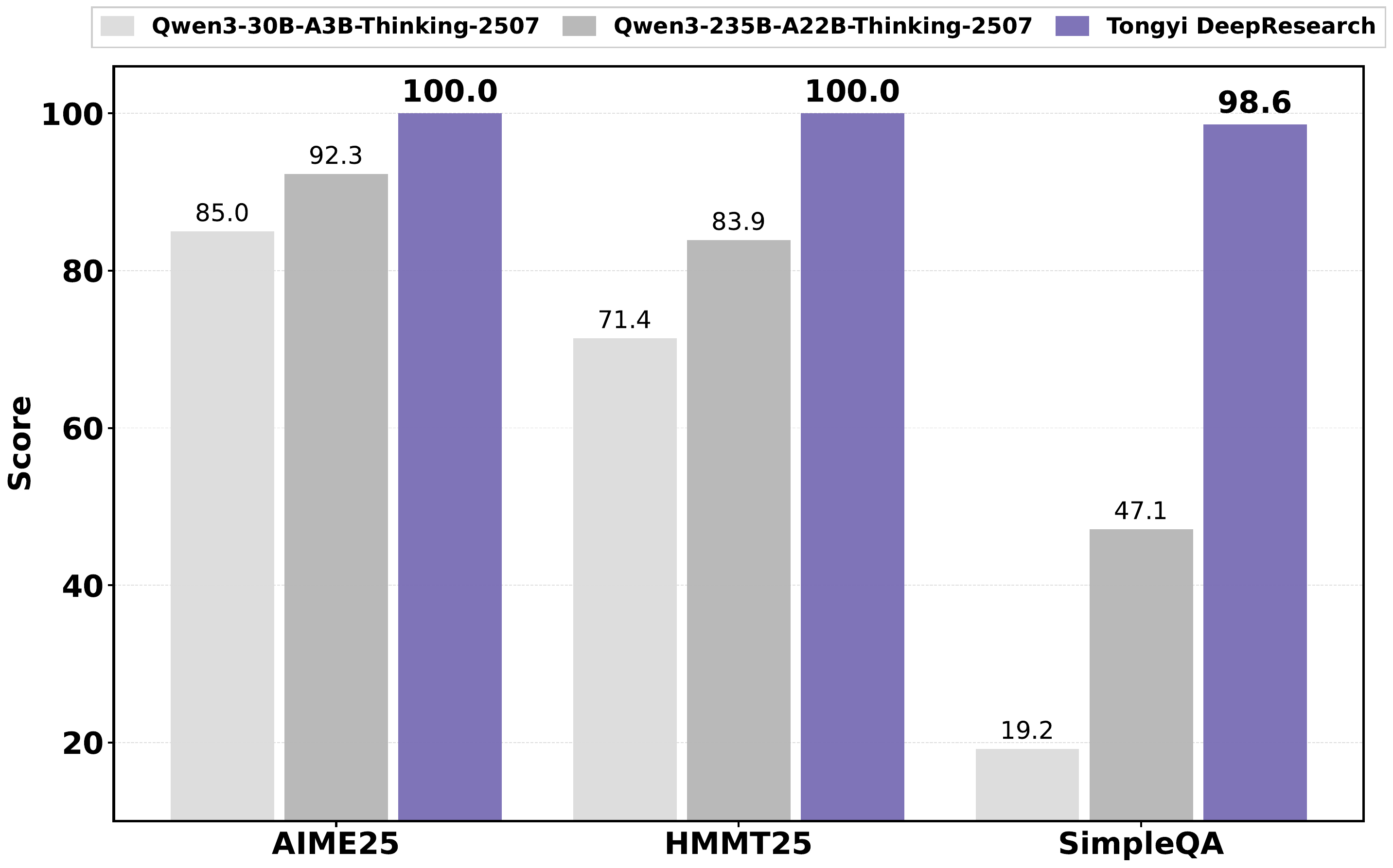}
 \caption{Performance on general benchmarks.}
\label{fig:general}
\end{wrapfigure}

\textbf{Performance on General Benchmark.}
We evaluate three general benchmarks, AIME25, HMMT25 and SimpleQA~\citep{simpleqa}.
The results are shown in Figure~\ref{fig:general}.
Experimental results demonstrate that Tongyi DeepResearch achieves substantial improvements over the base model, which relies solely on reasoning without any tool use.
On one hand, the system can retrieve external information via search, which proves particularly effective for knowledge-intensive benchmarks, and on the other, Python Interpreter enables it to enhance performance on mathematical reasoning tasks through native computational support.
Looking ahead, model training increasingly converges with agent training, solving paradigms evolve toward agentic architectures that integrate tool invocation and environment interaction, reflecting a more human-like problem-solving process.

\section{Discussion}
\subsection{Limitations}
We acknowledge several limitations in our current work:
First, the current 128K context length remains insufficient for handling the most complex long-horizon tasks, motivating further exploration of extended context windows or more advanced context management mechanisms~\citep{qiao2025webresearcher, wu2025resum}.
Second, we have not yet released a larger-scale model. Although the smaller-sized model already demonstrates strong performance, a larger model is currently in progress.
Third, we are continuously improving report generation fidelity and optimizing for user preferences to ensure more faithful, useful, and preference-aligned outputs~\citep{li2025webweaver}.
Fourth, we aim to improve the efficiency of our reinforcement learning framework by exploring techniques such as partial rollouts, which will require addressing off-policy training challenges, including distributional shift.
Finally, our current Deep Research training focuses on specific prompt instructions and predefined tool sets. 
We plan to enhance its robustness and extend the framework from Deep Research to broader agentic tool use scenarios.

\subsection{Model Scale}
We believe that training agentic capabilities on relatively small models is highly valuable~\citep{belcak2025small}.
Smaller models are inherently more efficient to deploy on edge devices, broaden accessibility across diverse real-world scenarios, and deliver faster, more responsive interactions.
This direction aligns with the broader goal of making autonomous research agents both powerful and practically deployable.

\subsection{What's Next}
We have a long-standing commitment to advancing research and development in deep research agents.
The Tongyi DeepResearch represents a significant step toward AI systems capable of autonomously transforming information into insight.
We advocate for open-source models with emergent agency, which are essential for democratizing agentic intelligence and deepening our fundamental understanding of how agency can emerge and scale in open systems.
Looking ahead, we aim to evolve from domain-specific agents to general-purpose agents, which are capable of reasoning, planning, and acting autonomously across diverse domains with minimal human supervision.
To achieve this, we are developing the \textbf{next-generation agent foundation model}, a unified model designed to endow AI systems with scalable reasoning, memory, and autonomy, enabling them to operate as truly general agents.
We believe it will empower individuals and organizations to reach new heights of productivity and innovation.

\section{Conclusion}
We introduced Tongyi DeepResearch, an open-source deep research agent that unifies agentic mid-training and post-training into a scalable, end-to-end paradigm.
Through automated data synthesis and stage-specific environments, the model learns to plan, search, reason, and synthesize information autonomously.
Despite its efficiency, activating only 3.3B parameters, Tongyi DeepResearch achieves state-of-the-art results on multiple deep research benchmarks, surpassing strong proprietary systems.
This work establishes a foundation for open, reproducible research into autonomous AI agents and marks a step toward more general, self-improving intelligence.
\clearpage
\section*{Contributions}
\label{sec:contribution}

The names are listed in alphabetical order by first name.

\textbf{Project Leader} \\
Yong Jiang

\textbf{Core Contributors} \\
Baixuan Li, Bo Zhang, Dingchu Zhang, Fei Huang, Guangyu Li, Guoxin Chen, Huifeng Yin, Jialong Wu, Jingren Zhou, Kuan Li, Liangcai Su, Litu Ou, Liwen Zhang, Pengjun Xie, Rui Ye, Wenbiao Yin, Xinmiao Yu, Xinyu Wang, Xixi Wu, Xuanzhong Chen, Yida Zhao, Zhen Zhang, Zhengwei Tao, Zhongwang Zhang, Zile Qiao

\textbf{Contributors} \\
Chenxi Wang, Donglei Yu, Gang Fu, Haiyang Shen, Jiayin Yang, Jun Lin, Junkai Zhang, Kui Zeng, Li Yang, Hailong Yin, Maojia Song, Ming Yan, Minpeng Liao, Peng Xia, Qian Xiao, Rui Min, Ruixue Ding, Runnan Fang, Shaowei Chen, Shen Huang, Shihang Wang, Shihao Cai, Weizhou Shen, Xiaobin Wang, Xin Guan, Xinyu Geng, Yingcheng Shi, Yuning Wu, Zhuo Chen, Zijian Li

\clearpage
\bibliography{biblio}
\bibliographystyle{colm2024_conference}

\appendix
\section{Rollout Details}

\begin{tcolorbox}[title=System Prompt]
You are a deep research assistant. Your core function is to conduct thorough, multi-source investigations into any topic. You must handle both broad, open-domain inquiries and queries within specialized academic fields. For every request, synthesize information from credible, diverse sources to deliver a comprehensive, accurate, and objective response. When you have gathered sufficient information and are ready to provide the definitive response, you must enclose the entire final answer within \textcolor{black}{\textbf{<answer></answer>}} tags. \\

\# Tools
\\

You may call one or more functions to assist with the user query.

You are provided with function signatures within \textcolor{black}{\textbf{<tools></tools>}} XML tags:\\
\textcolor{black}{\textbf{<tools>}}\\
\{"type": "function", "function": \{"name": "\textcolor{purple1}{\textbf{search}}", "description": "Perform Google web searches then returns a string of the top search results. Accepts multiple queries.", "parameters": \{"type": "object", "properties": \{"query": \{"type": "array", "items": \{"type": "string", "description": "The search query."\}, "minItems": 1, "description": "The list of search queries."\}\}, "required": ["query"]\}\}\}\\
\{"type": "function", "function": \{"name": "\textcolor{purple1}{\textbf{visit}}", "description": "Visit webpage(s) and return the summary of the content.", "parameters": \{"type": "object", "properties": \{"url": \{"type": "array", "items": \{"type": "string"\}, "description": "The URL(s) of the webpage(s) to visit. Can be a single URL or an array of URLs."\}, "goal": \{"type": "string", "description": "The specific information goal for visiting webpage(s)."\}\}, "required": ["url", "goal"]\}\}\}\\
\{"type": "function", "function": \{"name": "\textcolor{purple1}{\textbf{PythonInterpreter}}", "description": "Executes Python code in a sandboxed environment. To use this tool, you must follow this format:\\
1. The 'arguments' JSON object must be empty: \{\}.\\
2. The Python code to be executed must be placed immediately after the JSON block, enclosed within \textcolor{black}{\textbf{<code>}} and \textcolor{black}{\textbf{</code>}} tags.
\\\\
IMPORTANT: Any output you want to see MUST be printed to standard output using the print() function.
\\
\\
Example of a correct call:
<tool\_call>
\{"name": "PythonInterpreter", "arguments": \{\}\} \\
<code>
import numpy as np
\# Your code here
print(f"The result is: {np.mean([1,2,3])}")
</code>
</tool\_call>", "parameters": \{"type": "object", "properties": \{\}, "required": []\}\}\}\\
\{"type": "function", "function": \{"name": "\textcolor{purple1}{\textbf{google\_scholar}}", "description": "Leverage Google Scholar to retrieve relevant information from academic publications. Accepts multiple queries. This tool will also return results from google search", "parameters": \{"type": "object", "properties": \{"query": \{"type": "array", "items": \{"type": "string", "description": "The search query."\}, "minItems": 1, "description": "The list of search queries for Google Scholar."\}\}, "required": ["query"]\}\}\}
\{"type": "function", "function": \{"name": "\textcolor{purple1}{\textbf{parse\_file}}", "description": "This is a tool that can be used to parse multiple user uploaded local files such as PDF, DOCX, PPTX, TXT, CSV, XLSX, DOC, ZIP, MP4, MP3.", "parameters": \{"type": "object", "properties": \{"files": \{"type": "array", "items": \{"type": "string"\}, "description": "The file name of the user uploaded local files to be parsed."\}\}, "required": ["files"]\}\}\}\\
\textcolor{black}{\textbf{</tools>}}

For each function call, return a json object with function name and arguments within \textcolor{black}{\textbf{<tool\_call></tool\_call>}} XML tags:
\textcolor{black}{\textbf{<tool\_call>}}
\{"name": <function-name>, "arguments": <args-json-object>\}
\textcolor{black}{\textbf{</tool\_call>}}
\\
Current date: 
\end{tcolorbox}

The above constitutes the system prompt of our ReAct rollout.

\section{Evaluation Details}
\label{app:eval}
For GAIA (full validation set; 166 examples) and WebWalkerQA (680 examples), we adopt \texttt{Qwen2.5-72B-Instruct} as the judging model, following \cite{Li2025webthinker}.
For xbench-DeepSearch (100 examples) and xbench-DeepSearch-2510 (100 examples), we adopt \texttt{Gemini-2.0-Flash-001} as the judge model. 
For BrowseComp (1,266 examples) and BrowseComp-ZH (289 examples), we employ \texttt{GPT-4o-2024-08-06} as the judge model.
For Humanity’s Last Exam, we evaluate the 2,154 text-only questions following~\cite{chai2025scimaster}. 
The evaluation prompt follows the official protocol, with the \texttt{o3-mini} serving as the evaluator.
The evaluation prompt for these benchmarks is kept consistent with that described in the original paper to ensure alignment and reproducibility.
The evaluation prompts used for each benchmark is provided in detail on our GitHub repository\footnote{\url{https://github.com/Alibaba-NLP/DeepResearch/tree/main/evaluation}}.

For general benchmarks, we adopt different evaluation strategies based on task type.
For mathematical problems, since our system outputs a detailed report and datasets such as AIME25 (30 examples) and HMMT25 (30 examples) are relatively small in scale, we employ manual evaluation to ensure accuracy and fairness.
For knowledge-based problems, we utilize the official evaluation script of SimpleQA (4,326 examples) to maintain consistency with established benchmarks.

\section{Post-training Synthetic Data Case}
\label{app:data_case}

\begin{center}
\framebox{%
\begin{minipage}{15cm}
\ttfamily\footnotesize
{\color{blue}Question:}\\
A military officer, who also served as governor in a western North American territory, commanded a mounted infantry unit during a period of significant mineral discovery in the region. His official report on the discovery prompted the minting of a special commemorative coin in a certain year in the mid-19th century. During that same year, the unit he commanded was involved in a military conflict against a neighboring country. Just over a decade later, this unit was officially redesignated and would be assigned to a new division in the early 1920s. In the 1930s, this redesignated regiment was involved in an organizational swap. Which other regiment was it exchanged for?\\
{\color{purple}Answer:}\\
\textbf{12th Cavalry Regiment}
\end{minipage}
}
\end{center}

\begin{center}
\framebox{%
\begin{minipage}{15cm}
\ttfamily\footnotesize
{\color{blue}Question:}\\
An 18th-century travelogue, later adapted for a radio series, describes a port town in southeastern England as notable for its rampant illicit trade. This town was also the home of a 16th-century gentleman whose murder led to his wife's execution. Centuries later, another resident of the same town was granted letters patent providing special commercial privileges in a particular year of the early 19th century. During that same year, a collector, whose large collection of manuscript poems was later auctioned, secured a patent for a method of grinding inks. In that year, a patent of nobility was issued to a German family; what is the German term for the princely status it conferred?\\
{\color{purple}Answer:}\\
\textbf{Fürstenstand}
\end{minipage}
}
\end{center}

\begin{center}
\framebox{%
\begin{minipage}{15cm}
\ttfamily\footnotesize
{\color{blue}Question:}\\
In trisilylamine (N(SiH$_3$)$_3$), the Si-N bond length is 1.736 Å. Substituting one silyl group with methyl to form (CH$_3$)N(SiH$_3$)$_2$ elongates the Si-N bond to 1.752 Å. Calculate the percentage increase in bond length due to diminished hyperconjugation, and identify which specific orbital interaction weakens most significantly. Use covalent radii: Si=1.11 Å, N=0.70 Å, C=0.77 Å.\\
{\color{purple}Answer:}\\
\textbf{n $\rightarrow$ $\sigma^*_{Si-C}$}
\end{minipage}
}
\end{center}

The first two cases above are synthetically generated high-quality, high-uncertainty, superhuman question–answer pairs, examples of a caliber that is exceptionally difficult to produce via human annotation.
The third case represents a PhD-level research question, demanding deep domain expertise, multi-step reasoning.

\section{Environment Details}
\label{app:env}
We utilize five tools for Tongyi DeepResearch, namely Search, Visit, Python Interpreter, Google Scholar, and File Parser\footnote{Since our system relies on several internal APIs and fallback strategies (as described in Section~\ref{sec:rl}), we provide alternative open implementations in our open-source GitHub repository to facilitate public use. 
We have verified through extensive testing that these substitutions can faithfully reproduce our results.}:
\begin{itemize}[leftmargin=1.5em, itemsep=2pt, topsep=1pt, parsep=0pt]
    \item \textbf{Search} leverages the Google search engine for information retrieval. 
    The tool accepts a list of one or more search queries to be executed concurrently. 
    For each query, it returns the top-10 ranked results, with each result comprising a title, a descriptive snippet, and its corresponding URL.
    \item \textbf{Visit} is designed for targeted information extraction from web pages.
    The tool takes as input a set of web pages, where each page is paired with a dedicated information-seeking goal. 
    The process begins by employing Jina~\citep{jina} to parse the full content of a given web page.
    Subsequently, a summary model processes this content to extract only the information pertinent to that page’s specific goal.
    \item \textbf{Python Interpreter} is used to execute Python code within a sandboxed environment. 
    The input is a string of Python code, which must be enclosed within \textit{<code>} tags for proper execution. The tool runs the provided code and captures its standard output; therefore, any results or values intended to be seen must be explicitly passed to the \textit{print()} function. This capability enables dynamic computation, data manipulation, and the use of various Python libraries in a secure and isolated manner.
    \item \textbf{Google Scholar} is used to retrieve information from academic publications. The input consists of a list of one or more search queries, allowing for multiple, distinct searches within a single tool call. The tool leverages the Google Scholar search engine to execute each query and gather relevant scholarly literature, such as articles, papers, and citations.
    \item  \textbf{File Parser} answers user queries by analyzing a mix of documents, web pages, and multimedia files (\textit{e.g.}, PDF, DOCX, MP4) from local or URL sources. It works in two steps: first, it converts all input into plain text, transcribing audio/video content when necessary. Second, a summary model reads this unified text to generate a direct answer to the user’s question
\end{itemize}

\end{document}